\documentclass[runningheads]{llncs}
\usepackage{graphicx}
\begin{document}
\title{Cyrus 2D Simulation Team Description Paper 2016}
%
%

\author{Nader Zare\inst{3} \and
Ashkan Keshavarzi\inst{2} \and
Seyed Ehsan Beheshtian\inst{1} \and
Hadi Mowla\inst{1} \and
Aryan Akbarpour\inst{4} \and
Hossein Jafari\inst{1} \and
Keyvan Arab Baraghi\inst{1} \and
Mohammad Amin Zarifi\inst{1} \and
Reza Javidan\inst{1}
}
\authorrunning{N. Zare et al.}
%
\institute{
Shiraz University of Technology, Shiraz, Iran\\
\and
Tehran University, Tehran, Iran\\
\and
Khajeh Nasir Toosi University of Technology, Tehran, Iran\\
\and
Atomic Energy High School\\
\email{nader.zare@outlook.com}
\email{sebeheshtian@gmail.com}\\
\email{hadi.mowla@outlook.com}
\email{ho3einjafari92@gmail.com}\\
\email{keivan\_arab762@yahoo.com}
\email{Keshavarzi.a@ut.ac.ir}\\
\email{aryan.ak99@yahoo.com}
}
\maketitle 
\begin{abstract}
This description includes some explanation about algorithms and also algorithms that are being implemented by Cyrus team members. The objectives of this description is to express a brief explanation about shoot, block, mark and defensive decision will be given. It also explained about the parts that has been implemented. The base code that Cyrus used is agent 3.11. 

\keywords{RoboCup \and Soccer Simulation 2D \and Shoot.}
\end{abstract}
\section{Introduction}
Cyrus robotic team members founded this team five years ago with the goal of student scientific improvement in Artificial Intelligence and multi agent field. At first members of this team were bachelors of Information Technology school of Shiraz University of Technology and presently two of them are studding master degree in Tehran University and Khajeh Nasir University. This team has got permission to participate in world competition years 2013 to 2015 and could get 8th, 5th and 9th rank sequentially Also in 2014 achieved First place of Iran Open competition, First place in Kordestan 2013 and First place in Fazasazan 2012. This team has also participated in IranOpen 2013, IranOpen 2012, Sharif Cup 2012, IranOpen 2011, Sama RoboCup 2011 and etc.

Presently Cyrus team members are trying to improve team source code with 
Artificial Intelligence and multi agent algorithms. In this description shoot behavior which is implemented by rough neural networks algorithm is explained and also by using a simulation software developed with centralized processing of different defensive decision algorithm are being tested. We selected an optimized method of defensive decision and explained it which isn’t implemented yet.

\section{Shoot}
In paper [1] implement the shoot behavior with use neural network have one hidden layer. Briefly, at first in this method goal area will divide to 28 points with equidistant. Then for creating dataset of agent’s opponent positions, use ball position as input and best target of 28 points as output of neural network will be used, also use the best target in each run to learn weights of neural network. Best target due to the location of the ball and the opponent will calculate in this way that first, the path of the ball towards the specific goal is simulated and determines whether it is possible that the ball takeover by the opponent in this path or not. To clarify this point, location of the ball in each cycle calculated according to the initial speed of the ball toward the goal and ground friction. If before the ball crossing the goal line, in none of the point opponent can’t reach that point then shoot reach the target. Otherwise shoot can’t reach the target. After determining the status of all targets, the longest period of consecutive targets that are caused the ball reach the goal area is chosen as the best period for shooting and the center of this period is chosen as the best target for shooting. The number of target to be selected as the main output of intended state. This issue is shown in Figure1.
\begin{figure}[]
 \centering
 \includegraphics[scale=0.35]{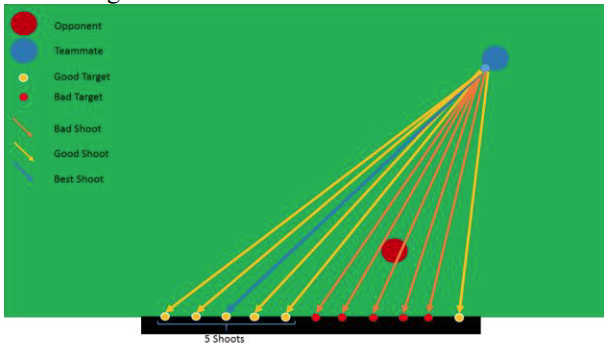}
 \caption{Target of shoots in paper [3] method.}
 \label{fig:Fig1}
\end{figure}
As mentioned previously, for learning neural network the status which consist of opponent’s location and location of the ball, to be considered as the input and number of the best target to be considered as the output of intended neural network. This approach has three problems:
\begin{enumerate}
 \item This approach finds the best target if only one player would have existed.
 \item The goal is obtained may not be the best goal in the dataset.
 \item The opponent’s action isn’t affected the algorithm, when this opponent has 
not seen yet.
\end{enumerate}
In the next section, the algorithm to find the best target is described.

\subsection{Cyrus shoot algorithm}
In the shoot algorithm implementation in Cyrus team, in the first step the opponents’ 
position and also ball position evaluation will be calculated for each shoot to specific 
target. Then the shoot with highest evaluation number is chosen as the best possible 
shoot to achieve the goal. 
In this algorithm, the first ball movement to the specified target is simulated, then 
opponent player’s movement based on the ball path will be simulate and time that 
need to opponent reach this path will be calculate to find best shoot base on 
evaluations. 
World model of soccer 2D simulation has uncertain and in observable environment. 
So we cannot tell for sure that shot is going to be scored or not. At first we will 
evaluate the targets with some parameters like: position of the opponent agents, 
opponent agent’s type, ball position and the maximum initial velocity of the ball. For 
evaluating the targets of the shoot we will use the above parameters and simulate the 
ball path to the target position as way ball position in each cycle to be clear. Then in 
this algorithm, the time is opponent need to reach the ball will be calculated.
\begin{figure}[]
 \centering
 \includegraphics[scale=0.35]{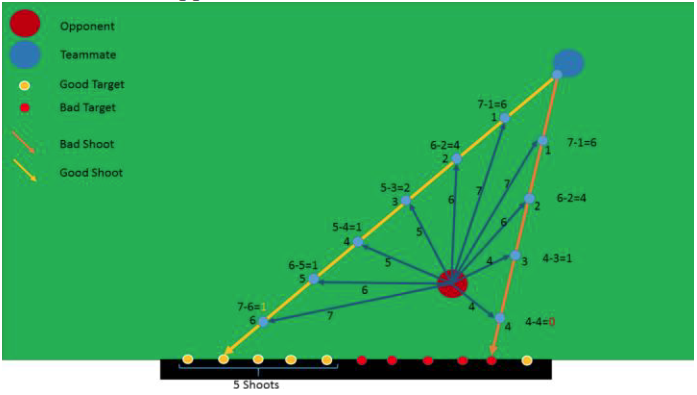}
 \caption{Shoots can be reach the targets and opponent reach time.}
 \label{fig:Fig1}
\end{figure}
In first step as described in previous sections, this algorithm is scoring all point of 
the path from ball to the target and output of this algorithm is shown in figure 2.

\subsection{Using neural networks}
The goal of using neural networks in the Cyrus shoot algorithm is to obtain evaluation 
of each shoot to goal based on one player. In the Cyrus team because of calculation 
time in the soccer 2D simulation environment to simulate opponent reach time to the 
shoot path is much expensive and we use neural network. Presently we used opponent 
player’s positions, target position and ball position as input of our neural networks, 
and also first ball speed is considered 3.0 but in future before world competition we 
want to add body angle, opponent player’s first speed and shoot speed to learn neural 
networks. 
Also in soccer 2D environment, poscount has direct impact on shoot behavior and 
other behaviors in the soccer 2D simulation environments. To optimize neural 
networks output with use of poscount, multiplied with a certain factor and add to 
output of neural networks. Also this factor will be calculated by reinforcement 
learning in huge of games dataset.
\begin{figure}[]
 \centering
 \includegraphics[scale=0.35]{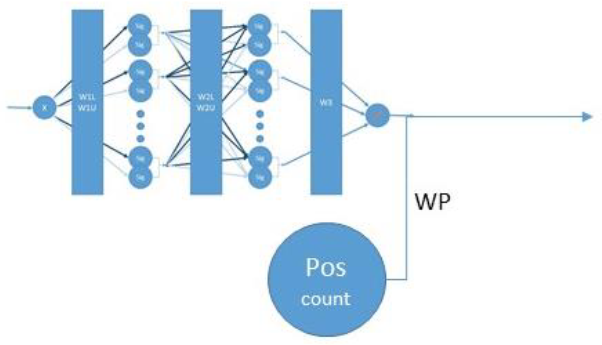}
 \caption{Add poscount to neural network output.}
 \label{fig:Fig1}
\end{figure}

\subsection{Comparison}
In next sections and figure 7 a compare between shoot implemented in agent2D [2] 
version 3.1.1, paper [1] and Cyrus implemented shoot is shown:

\begin{table}
\centering
\caption{compare shoot implementation.}\label{tab1}
\begin{tabular}{|l|l|l|l|}
\hline
Algorithms: & Agent2D & Paper[1] & CYRUS NN\\
\hline
Win Rate & 48.6\% & 55.3\% & 59.2\%\\
\hline
\end{tabular}
\end{table}

\subsection{Optimization using pattern recognition}
For optimizing data from previous method first we turn our data into patterns and then 
we use our new data set as our neural networks inputs. In this method our previous 
data consisting of ball position, opponent players and target position are turned into 
ball to target distance, ball to opponent distance and the angle between ball-target and 
ball-opponent vectors. In this method symmetry property of neural networks is kept 
and better results are obtained. In figure 8 we will see errors has been obtained by 
learning our two sets of data.
\begin{figure}[]
 \centering
 \includegraphics[scale=0.35]{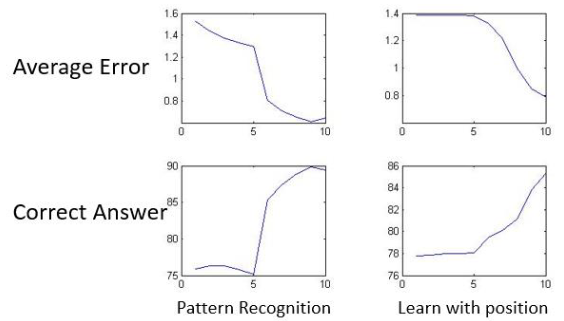}
 \caption{Errors obtained by learning our two sets of data.}
 \label{fig:Fig1}
\end{figure}

\section{Future Works}
Cyrus team wants to use rough deep neural networks for offensive decision making. 
In this method we want to use Helios [4], WrightEagle [5], Gliders [6] and Oxsy [7] 
binaries to use player positions as input and player’s behavior as output of neural 
networks. After learning the neural networks, we want to use reinforcement algorithm 
to optimize this neural networks. For optimizing deep neural networks by 
reinforcement algorithm we want to divide field into different parts and according to 
ball reaching to those places based on the dangerous and importance of each region 
we will reward each player and then with help of these rewards we will update 
weights of our neural networks. This method can be used in three different ways:
\begin{itemize}
 \item Learning other teams’ offensive behavior and using it for defensive decisions.
 \item Each of our players can predict its teammate’s behavior.
 \item We can improve our team’s offensive decisions by Re-Enforcement Learning.
\end{itemize}
We also want to use generalized kalman filter’s algorithm online to predict team-
mate and opponent positions. In this algorithm we use player position in each cycle as 
input of our neural networks and get its next position and by observing its next real 
position. We will compare our prediction to reality and learn our neural networks 
while running. In the live game we want to reduce poscount impact by using this 
method and also in offensive time by using chain-action we can improve the ball 
holder’s future vision.

\end{document}